\documentclass[review,onefignum,onetabnum]{siamonline181217}

\usepackage{amssymb}
\usepackage{multirow}
\usepackage{subfigure}
\usepackage{graphicx}
\usepackage{placeins} 
\usepackage{bm}

\usepackage{lipsum}
\usepackage{amsfonts}
\usepackage{graphicx}
\usepackage{epstopdf}
\usepackage{algorithmic}
\ifpdf
  \DeclareGraphicsExtensions{.eps,.pdf,.png,.jpg}
\else
  \DeclareGraphicsExtensions{.eps}
\fi

\usepackage{enumitem}
\setlist[enumerate]{leftmargin=.5in}
\setlist[itemize]{leftmargin=.5in}

\newcommand{\creflastconjunction}{, and~}

\newsiamremark{remark}{Remark}
\newsiamremark{hypothesis}{Hypothesis}
\crefname{hypothesis}{Hypothesis}{Hypotheses}
\newsiamthm{claim}{Claim}

\headers{A Regularized CNN for Semantic Image Segmentation}{Fan JIA, Jun LIU, and Xue-cheng TAI}

\title{A Regularized Convolutional Neural Network for Semantic Image Segmentation
}

\author{Fan JIA\thanks{Department of Mathematics, Hong Kong Baptist University 
		(\email{jiafan@life.hkbu.edu.hk}, \email{xuechengtai@hkbu.edu.hk}).}
	\and Jun LIU\thanks{Laboratory of Mathematics and
		Complex Systems (Ministry of Education of China), School of Mathematical
		Sciences, Beijing Normal University 
		(\email{jliu@bnu.edu.cn}).}
	\and Xue-cheng TAI\footnotemark[1]}

\usepackage{amsopn}

\ifpdf
\hypersetup{
pdftitle={A Regularized Convolutional Neural Network for Semantic Image Segmentation},
pdfauthor={Fan JIA , Jun LIU, and Xue-cheng TAI}
}
\fi

\begin{document}

\maketitle

\begin{abstract}
	Convolutional neural networks (CNNs) show outstanding performance in many image processing problems, such as image recognition, object detection and image segmentation. Semantic segmentation is a very challenging task that requires recognizing, understanding what's in the image in pixel level. Though the state of the art has been greatly improved by CNNs, there is no explicit connections between prediction of neighbouring pixels. That is, spatial regularity of the segmented objects is still a problem for CNNs. In this paper, we propose a method to add spatial regularization to the segmented objects. In our method, the spatial regularization such as total variation (TV) can be easily integrated into CNN network. It can help CNN find a better local optimum and make the segmentation results more robust to noise. We apply our proposed method to Unet and Segnet, which are well established CNNs for image segmentation,  and test them on WBC, CamVid and SUN-RGBD datasets, respectively. The results show that the regularized networks not only could provide better segmentation results with regularization effect than the original ones but also have certain robustness to noise. 
\end{abstract}

\begin{keywords}
	CNN, Image Segmentation, Regularization.
\end{keywords}

\begin{AMS}
	68U10, 68T10, 65K10
\end{AMS}

\section{Introduction}
Convolutional Neural Networks (CNNs) \cite{lecun1998gradient} have achieved prominent performance in a series of image processing tasks, such as image classification \cite{he2016deep,krizhevsky2012imagenet,zeiler2014visualizing}, object detection \cite{erhan2014scalable,girshick2014rich,he2016deep,liu2016ssd,papandreou2015modeling,sermanet2013overfeat} and image segmentation \cite{badrinarayanan2015segnet,chen2018deeplab,long2015fully,ronneberger2015u,zhao2017pyramid}. CNNs which dwarf systems relying on hand-crafted features use millions of parameters to learn latent features from large scale training datasets. It is intractable to design a hand-crafted feature which allows to learn increasingly abstract data representations. However, CNNs could do it well \cite{zeiler2014visualizing}.

Image segmentation is a process of segmenting a digital image into different regions. It aims to simplify and/or change the representation of an image into something that is more meaningful and easier to analyze. Semantic image segmentation is a much more challenging segmentation task. It requires understanding the image in pixel level. More precisely, it is a process of assigning a label to every pixel in an image such that pixels with the some label share some common features \cite{barghout2004perceptual,shapiro2001computer}. 

\begin{figure}[tbhp]
	\centering
	\begin{subfigure}[clean image \label{f1a}]
		{
			\includegraphics[width=0.22\linewidth]{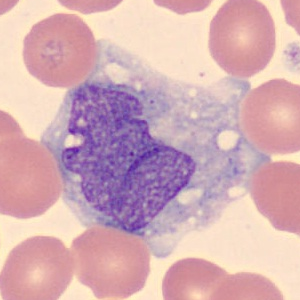}
		}
	\end{subfigure}
	\begin{subfigure}[Unet \label{f1b}]
		{
			\includegraphics[width=0.22\linewidth]{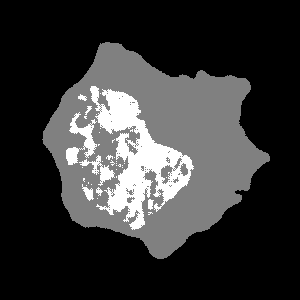}
		}
	\end{subfigure}
	\begin{subfigure}[Unet+TV \label{f1c}]
		{
			\includegraphics[width=0.22\linewidth]{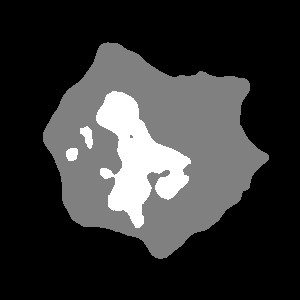}
		}
	\end{subfigure}
	\begin{subfigure}[RUnet \label{f1d}]
		{
			\includegraphics[width=0.22\linewidth]{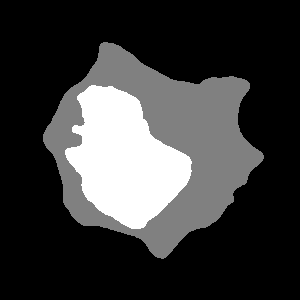}
		}
	\end{subfigure}
	\begin{subfigure}[noisy image \label{f1e}]
		{
			\includegraphics[width=0.22\linewidth]{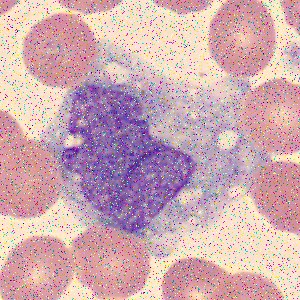}
		}
	\end{subfigure}
	\begin{subfigure}[Unet \label{f1f}]
		{
			\includegraphics[width=0.22\linewidth]{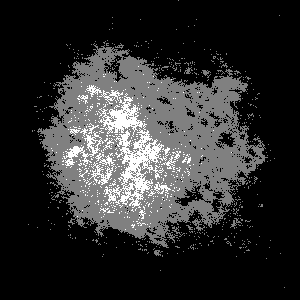}
		}
	\end{subfigure}
	\begin{subfigure}[Unet+TV \label{f1g}]
		{
			\includegraphics[width=0.22\linewidth]{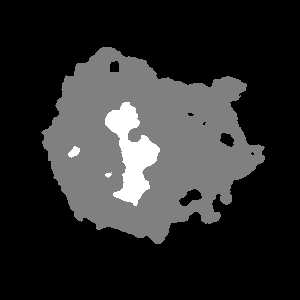}
		}
	\end{subfigure}
	\begin{subfigure}[RUnet \label{f1h}]
		{
			\includegraphics[width=0.22\linewidth]{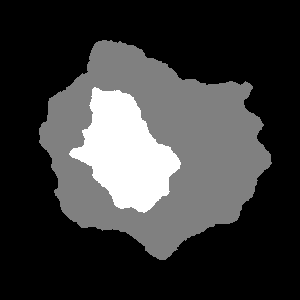}
		}
	\end{subfigure}
	\caption{An example of segmentation results by performing the original Unet \cite{ronneberger2015u} and our proposed regularized Unet (RUnet) on WBC Dataset\cite{Zheng2018}. When adding noise to image, the segmentation of nucleus by Unet becomes messy (\Cref{f1f}). If we add post-processing to the prediction of Unet (Unet+TV),  the result is still not desirable (\Cref{f1c},\Cref{f1g}). However, our proposed RUnet could provide smooth segmentation result (\Cref{f1d},\Cref{f1h}).}
	\label{fig:image_segmentation_problems}
\end{figure}

Currently, CNNs regard semantic image segmentation as a dense prediction problem. They predict the classification of each pixel independently. Though the convolution kernels are shareable and predictions to different pixels have implicit connection. There is no explicit connection when predicting adjacent pixels which is not accord with fact that classifications of pixels inside an object are interdependent. That is, spatial regularization is still missing. Though many efforts have been made and the accuracy has been continuously improved, CNNs fail to provide segmentation results with regularization effect. This is because the popular CNNs are usually continuous mappings which are composite mappings of continuous operators such as affine transformations and continuous activation functions (e.g. Soft-max, ReLU, Sigmoid). They can not provide spatial regularization for the segmented objects. As shown in \Cref{fig:image_segmentation_problems}, we trained the original Unet \cite{ronneberger2015u} and our RUnet on  White Blood Cell (WBC) Dataset \cite{Zheng2018}. To test the robustness of the two methods with respect to noise, we test the two trained network on image with added noise. The segmentation results of the original Unet \cite{ronneberger2015u} becomes much worse, though the noise level is not high. Nevertheless, RUnet can still achieve good segmentation result.

Image restoration such as denoising and deblurring are the most fundamental tasks in image processing. It is important but difficult to preserve image structures
(such as edges) in image restoration \cite{wu2010augmented}. Total Variation (TV) method shows good performance when handling minimizing problems in image restoration \cite{chambolle2004algorithm,chambolle1997image,rudin1992nonlinear}, since it can preserve discontinuity. We shall propose a way to add spatial TV regularization to CNNs.

Our essential idea is to add spatial regularization to activation functions. In this paper, we focus on applying spatial regularization to softmax. The same technique can also be applied to other activation functions. This gives us CNNs with spatial regularity.

We apply our proposed method to Unet \cite{ronneberger2015u}, Segnet \cite{badrinarayanan2015segnet} and evaluate their performance on WBC Image Dataset \cite{Zheng2018}, CamVid Dataset\cite{johnson2016driving} and SUN-RGBD Dataset\cite{song2015sun}. Unet is well known for its outstanding performance on biomedical image segmentation. It achieved the first place in ISBI cell tracking challenges 2015 leading other methods by a large margin. Segnet achieves better performance on real world scenes such as CamVid and SUN-RGBD datasets than DeepLab-LargeFOV \cite{chen2014semantic}, FCN \cite{long2015fully}, DeconvNet \cite{noh2015learning}. WBC Dataset consists of color cell images, which are collected from the CellaVision blog. The cell images allow us to observe distinct object details. CamVid Dataset and SUN-RGBD Dataset are much more complex datasets which consist of real world road scenes and indoor scenes, respectively. They are chosen as benchmark of Segnet.

Unet with regularized softmax (RUnet) and Segnet with regularized softmax(RSegnet) achieve better performance than original Unet and Segnet on testing datasets. The segmentation result is spatially regularized and robust. Our approach gives a promising direction for semantic segmentation tasks, which may benefit a series of CNN-based image processing tasks. Our main contributions are in the following:

\begin{itemize}
	\item[$\bullet$]  We propose a framework to integrate the traditional variational regularization method into deep convolutional neural networks. In this work, we present it for the softmax activation function. The same idea can be applied to other activation functions. It is known that spatial regularization is important for image and vision problems. So far, it is still missing to have good spatial regularization effects for these applications.
	
	\item[$\bullet$] When spatial regularization is added to CNNs, one essential difficulty is how to find a simple and clear way to calculate the gradient decent direction for general loss functions. We first give the general formula of computing gradients when integrate total variation to CNN. Then we propose an efficient method which needs very little modifications to existing CNNs and their numerical implementations, but has very visible regularization effects with much better robustness to noise and good improved accuracy.  
	
	\item[$\bullet$] We give experiments of our proposed method on two CNNs for image segmentation, i.e.  Unet and Segnet. By testing them on three datasets, it is numerically verified that the new method could produce smoother objects and has better robustness to noise.
\end{itemize}

The paper is organized as follows. In \Cref{sec:Related}, we give brief descriptions to related work , general neural network for semantic image segmentation and total variation, respectively. Our proposed method is given in \Cref{sec:proposed}. In this section, we apply our proposed method to softmax layer and give the general formulas for forward propagation, backward propagation. Some implementation details are also illustrated here. The experimental results are described  in \Cref{sec:results}, and the conclusions follow in \Cref{sec:conclusions}.

\section{Related Work}
\label{sec:Related}

Semantic segmentation task has long been an attractive topic in image processing. In early years, systems relying on hand-crafted features are combined with classifiers, such as Boosting \cite{ladicky2010and,sturgess2009combining},  Random Forests \cite{brostow2008segmentation,shotton2008semantic}, or Support Vector Machines \cite{brostow2008segmentation}. These methods often use region based method to predict the probability of each pixel. However, the choice of hand-crafted features can be very crucial, the performance of same feature can vary much when applied to different kinds of datasets. Meanwhile, the performance of such systems is compromised by insufficient feature representation ability. 

R-CNN \cite{girshick2014rich} and SDS \cite{hariharan2014simultaneous} use CNN as feature extractor which followed by final refinement step to help improve segmentation. Nevertheless, for pixel-wise semantic segmentation problems, the region based approach becomes bottleneck.

FCN \cite{long2015fully} is a successful attempt training end-to-end, pixel-to-pixel convolutional network on semantic segmentation.  It achieved 30\% relative improvement compared with previous best PASCAL VOC 11/12 test results. 

After FCN, a series of CNNs come out to improve the segmentation performance, such as Unet \cite{ronneberger2015u}, PSPNet \cite{zhao2017pyramid}, Segnet \cite{badrinarayanan2015segnet} and Deeplab \cite{chen2018deeplab}. Unet fuses high level feature map with low level feature map and shows prominent performance in medical image processing. PSPNet exploits the capability of global context information by different-region-based context aggregation. Segnet uses the max pooling indices
to upsample (without learning) the feature map(s) and convolves with a trainable decoder filter bank. Deeplab applies atrous convolution for dense feature extraction and enlarge the field-of-view. 

Some CNNs try to boost their ability to capture fine details by employing a fully-connected Conditional Random Field (CRF) \cite{krahenbuhl2011efficient}. However they didn't fully integrate CRF into CNN, thus CRF didn't contribute to updating the weights. Technique has also been proposed to regularize the parameter set to align results to edges \cite{ochs2016techniques}. Nevertheless, no one has tried to regularize segmentation results by adding spatial regularization to activation functions. Next, we will review the general CNN for semantic image segmentation and TV regularization.

\subsection{General Neural Network for Semantic Image Segmentation}
Let $  v \in\mathbb{R}^{N_1N_2}$ be a column vector by 
stacking the columns of image with size $N_1\times N_2$.
Taking $  v$ as an input of
a pixel-wise segmentation neural network. Mathematically, this network can be written as a parameterized nonlinear operator $\mathcal{N}_{ \Theta}$ defined by $  v^K=\mathcal{N}_{ \Theta}(  v)$. The output $  v^K$ of the network is given by the following recursive connections
\begin{equation}\label{eq:nn}
\left\{
\begin{array}{rl}
v^0=&  v,\\
v^{k}=&\mathcal{A}^{k}(\mathcal{T}_{  \Theta^{k-1}}(  v^{k-1})), k=1, \ldots, K, \\
\end{array}
\right.
\end{equation} 
where $\mathcal{A}^{k}$ is an activation function (e.g. sigmoid, softmax, ReLU) or sampling (e.g. downsampling, upsampling, dilated convolution) operator or their compositions and 
$\mathcal{T}_{  \Theta^{k-1}}$ is often chosen as an affine transformation with the representation $\mathcal{T}_{  \Theta^{k-1}}(  v)= \mathcal{  W}^{k-1}   v +   b^{k-1}$, in which $\mathcal{  W}^{k-1},   b^{k-1}$ are linear operator (e.g. convolution) and translation, respectively. The parameter set is
$ \Theta=\{ \Theta^k=(\mathcal{  W}^{k},   b^k)| k=0,  \ldots, K-1\}$.
The output of this network $  v^K\in\{0,1\}^{C\times N_1N_2}$ should be a binary classification matrix whose $c$-th column is a binary characteristic vector  for $c$-th class, $c=1,2, \ldots,C$. 
By giving $M$ images $\mathcal{V}=(v_1, v_2, \ldots,$  $ v_M)\in\mathbb{R}^{M\times N_1 N_2}$ and their ground truth segmentation $ \mathcal{U }=stack(\mathcal{U}_1,\mathcal{U}_2,$ $ \ldots,\mathcal{U}_M) \in\{0,1\}^{M\times C \times N_1N_2}$ with $\mathcal{U}_m \in\{0,1\}^{C\times N_1N_2}$, the training process is to learn a parameter set $ \Theta$  which minimizes a loss functional $ \mathcal{L}(\mathcal{N}_{ \Theta}(\mathcal{V}), \mathcal{U})$, namely
\begin{equation}
\Theta^*=\underset{\Theta}{\arg\min}~~\mathcal{L}(\mathcal{N}_{\Theta}(\mathcal{ V}), \mathcal{ U}).
\end{equation}

In many references, the loss functions are set to be the cross entropy which is given by 
\begin{equation}
\mathcal{L}(\mathcal{N}_{ \Theta}(\mathcal{  V}), \mathcal{  U})=-\frac{1}{M}\sum_{m=1}^M <\mathcal{U}_m, \log \mathcal{N}_{ \Theta}(  v_m)>.
\end{equation}

The algorithm of learning is a gradient descent method:
\begin{equation}
( \Theta^{k})^{step}=( \Theta^{k})^{step-1}-\tau_\Theta\frac{\partial \mathcal{L}}{\partial  \Theta^{k}}\Big|_{ \Theta^{k}=( \Theta^{k})^{step-1}},
\end{equation}
where $k=0, \ldots,K-1,step=1,2, \ldots$ is the iteration number and $\tau_\Theta$ is a time step or so called learning rate.
$\frac{\partial \mathcal{L}}{\partial  \Theta^{k}}$ can be calculated by backpropagation technique using chain rule.
Denoting $  o^k=\mathcal{T}_{  \Theta^{k-1}}(  v^{k-1})$, then
\Cref{eq:nn} becomes
\begin{equation}\label{eq:nn1}
\left\{
\begin{array}{rl}
v^{k}=&\mathcal{A}^{k}(  o^k),\\
o^k=&\mathcal{T}_{  \Theta^{k-1}}( v^{k-1}),
\end{array}
\right.
\end{equation}
where $k=1, \ldots, K$.
Let us write $\Delta^k=\frac{\partial \mathcal{L}}{\partial o^k}$, then the backpropagation scheme becomes 
\begin{equation}\label{bp}
\left\{
\begin{array}{rl}
\Delta^k&=\frac{\partial v^{k}}{\partial o^{k}}\cdot\frac{\partial o^{k+1}}{\partial v^{k}}\cdot\frac{\partial \mathcal{L}}{\partial o^{k+1}} \\
&=\frac{\partial \mathcal{A}^k}{\partial o^k}\cdot\frac{\partial \mathcal{T}_{ \Theta^{k}}}{\partial v^k}\cdot \Delta^{k+1}, \\
\frac{\partial \mathcal{L}}{\partial   \Theta^k}&=\frac{\partial o^{k+1}}{\partial  \Theta^k}\cdot\frac{\partial \mathcal{L}}{\partial o^{k+1}}=\frac{\partial \mathcal{T}_{ \Theta^{k}}}{\partial \Theta^k}\cdot\Delta^{k+1},
\end{array}
\right.
\end{equation}
where $k=0,1,\ldots,K-1$.

\subsection{Total Variation}

The total variation (TV) is proposed to produce piece-wise constants cartoon restorations in ROF model \cite{rudin1992nonlinear}.
TV can be written as
\begin{equation}
\begin{aligned} 
\text{TV}(u)=\int_{\Omega}|\triangledown u(x)| dx,
\end{aligned}
\end{equation}
where $\Omega$ is a bounded subset of $\mathbb{R}^2$, $u$ is a single channel image.
It has a dual formulation as
\begin{equation}
\begin{aligned}
\label{eq:TV_constrain}
\text{TV}(u)=&\underset{ \xi\in\mathbb{B}}{\sup}\left\{\int_{\Omega}u(x)div  \xi(x) dx \right\},
\end{aligned}
\end{equation}
where $\mathbb{B}=\{\xi\in C_0^1(\Omega;\mathbb{R} ^2)~|~ || \xi||_{\infty}=\underset{x \in\Omega}{\max}\{||  \xi(x)||_2\}\leq1\}$.

When $u$ has multi-channels, we  sum up the contributions of the separate channels, and the definition is given by the following:
\begin{equation}
\begin{aligned}
\label{eq:multi_TV}
\text{TV}(\bm u)=\sum_{i=c}^C\int_{\Omega}|\triangledown u_c(x)| dx 
\end{aligned}
\end{equation}
where $C$ is the number of channels. The dual formulation is given by:
\begin{equation}
\begin{aligned}
\label{eq:multi_TV_dual}
\text{TV}(\bm u)=&\underset{ \xi_1,\ldots,\xi_C\in\mathbb{B}}{\sup}\left\{\sum_{c=1}^C\int_{\Omega}u_c(x)div  \xi_c(x) dx \right\}.
\end{aligned}
\end{equation}  

For discrete TV and the related dual formulation, they have the similar expressions  \cite{chambolle2004algorithm}.

\section{Proposed Method}
\label{sec:proposed} 
\subsection{Intuition}
Usually, a CNN contains dozens of activation functions and softmax function is the most commonly used in the last layer. Softmax function is a function that takes as input a vector of C real numbers, and normalizes it into a probability distribution consisting of C probabilities.

In fact, softmax could be derived from a minimization problem. When given $\bm o\in\mathbb{R}^{C \times N_1N_2}$  as the input, $C$ is the number of classes, $N_1\times N_2$ is the image size, we want to find a corresponding output $\mathcal{\bm A}\in\mathbb{R}^{C \times N_1N_2}$ such that $\mathcal{\bm A}$ is the minimizer of the following  problem: 
\begin{equation}\label{softmax}
\begin{aligned}
&\min-<\mathcal{\bm A},\bm o> + <\mathcal{\bm A}, \log{\mathcal{\bm A}}>, \\
&s.t. \sum_{c}^{C}\mathcal{\bm A}_{ci} = 1, \forall i=1,\ldots,N_1N_2.
\end{aligned}
\end{equation}

Let $\mathcal{\bm A} = (\mathcal{\bm A}_1,\ldots,\mathcal{\bm A}_C), \mathcal{\bm A}_c\in\mathbb{R}^{ N_1N_2} $ for $c=1,2,\ldots C$. Some simple calculations can show that the minimizer of the above problem is: 
\begin{equation}
\mathcal{\hat{\bm A^*}}_{j}=\frac{\exp(\bm o_{j})}{\sum_{c=1}^C\exp(\bm o_{c})}, j=1,\ldots,C.
\end{equation}

$\mathcal{\hat{\bm A^*}}_{j}$ is the $j$-th class probability map of the input image. One can easily see that this is just the commonly used softmax activation function, i.e. 
\begin{equation}
\mathcal{\hat{\bm A^*}}=\text{Softmax}(\bm o)
\end{equation}

However, this function doesn't have any spatial regularization. Prediction of each pixel is independent of other pixels.

\subsection{Proposed Regularized Softmax Layer}
Inspired by the softmax variational problem \Cref{softmax},  we propose to replace the softmax function by the following regularized softmax:
\begin{equation}\label{eq:rnn}
\begin{array}{ll}
&\min-<\mathcal{\bm A},\bm o> + <\mathcal{\bm A}, \log{\mathcal{\bm A}}> + \lambda TV(\mathcal{\bm A}), \\
&s.t. \sum_{c}^{C}\mathcal{\bm A}_{ci} = 1, \forall i=1,\ldots,N_1N_2,
\end{array}
\end{equation}
where $\lambda$ is the regularization parameter which controls the regularization effect.
According to the definition of multi-channels' total variation \Cref{eq:multi_TV} and \Cref{eq:multi_TV_dual}, we have
\begin{equation}
\begin{aligned}
\label{eq:multi_A}
\text{TV}(\mathcal{\bm A})&=\sum_{c=1}^C\int_{\Omega}|\triangledown \mathcal{\bm A}_{c}(x)| dx = \underset{\xi_1,\ldots,\xi_c\in\mathbb{B}}{\sup}\left\{\sum_{c=1}^C\int_{\Omega}\mathcal{\bm A}_{c}(x)div \bm \xi_{c}(x) dx \right\} 
\end{aligned}
\end{equation}

Compared to the traditional neural network \Cref{eq:nn1}, the activation function $\mathcal{\bm A}$ is replaced by the solution of a TV regularized minimization problem. This is significantly different from the existing continuous neural network mappings.
Moreover, the problem in \Cref{eq:rnn} can be easily solved by primal-dual method:
\begin{equation}
\begin{aligned}
(\mathcal{\tilde{\bm A}}^*, \bm \eta^*)=&\arg\underset{\mathcal{\bm A}}{\min}~~\underset{ \bm \xi\in\mathbb{B}}{\max}
\{-<\mathcal{\bm A}, \bm o> + <\mathcal{\bm A},\log \mathcal{\bm A}>
+\lambda <\mathcal{\bm A}, div \bm \xi>\}, \\
& s.t. \sum_{c}^{C}\mathcal{\bm A}_{ci} = 1, \forall i=1,\ldots,N_1N_2. \\
\end{aligned}
\end{equation}

Similar to the Chambolle type projection algorithm \cite{chambolle2004algorithm} 
, the solutions of the above min-max problem satisfies the following relationship:
\begin{equation}
\label{eq:result}
\begin{array}{rl}
\mathcal{\tilde{\bm A}}^*_{j}&=\frac{\exp(\bm o_{j}-\lambda div{\bm \eta^*_{j}})}{\sum_{c=1}^C\exp(\bm o_{c} -\lambda div{\bm  \eta^*_{c}})}, j=1,\ldots,C.
\end{array}
\end{equation}

We can use the following primal-dual gradient algorithm to find the solution in an iterative way: 
\begin{equation}\label{eq:dualalgo}
\left\{
\begin{array}{ll}
\bm \xi^{t+1} &=\bm  \xi^t - \tau\lambda\triangledown \mathcal{\bm A}^{t} , \\
\bm \eta^{t+1} &= \mathcal{P}_\mathbb{B}(\bm \xi^{t+1}), \\
\mathcal{\bm A}^{t+1} &= \mathcal{S}(\bm o-\lambda div(\bm \eta^{t+1})), \\
\end{array}
\right.
\end{equation}
where $\mathcal{S}$ is the softmax operator, $t$ is the iteration number and $\tau$ is a time step, $\mathcal{P}_\mathbb{B}$ is a projection operator onto the convex set $\mathbb{B}$, given $\bm y =(y_1,y_2)=\bm \xi_{cj}$, $\mathcal{P}_\mathbb{B}(\bm y)$ is defined by
\begin{equation}\label{eq:projection}
\mathcal{P}_\mathbb{B}(\bm y) = 
\left\{
\begin{array}{ll}
\bm y, &\ if \ \|\bm y\|_2\le 1 \\
\frac{\bm y}{||\bm y||}, &\ if \ \|\bm y\|_2>1\\
\end{array}
\right.
\end{equation}

So $\mathcal{P}_\mathbb{B}(\bm \xi)$ refers to project every $\bm \xi_{cj}\in \bm \xi$ onto $\mathbb{B}$.
Mathematically, 
$\bm \eta^*=\underset{t\rightarrow +\infty}{\lim}\bm \xi^{t+1}$ and  
$\mathcal{\bm \tilde A}^* =  \underset{t\rightarrow +\infty}{\lim} \mathcal{\bm A}^{t+1} $. 
In real computation,
when the iteration \Cref{eq:dualalgo} converges, we can get $\mathcal{\tilde{\bm A}}^*$ and $\bm \eta^*$.

Thus, given $\bm o$ as the input of regularized softmax layer, we perform \Cref{eq:dualalgo} to obtain a convergent $\mathcal{\tilde{\bm A}}$ and $\bm \eta$, then the new regularized activation function has the following simple  expression:
\begin{equation}
\mathcal{\tilde{\bm A}}=\text{Softmax}(\bm o-\lambda div \bm \eta):=\mathcal{S}(\bm o-\lambda div \bm  \eta).
\end{equation}

\subsection{Regularized ReLU Layer}

Our proposed method could bring regularization effect to the segmentation results and the similar idea could be easily applied to other activation functions.

For example, the popular ReLU activation function is exactly the solution to the following minimization problem:
\begin{equation}
ReLU(\bm  o)=\underset{\mathcal{\bm A} \geqslant 0}{\arg\min}\left\{\frac{1}{2}|| \bm o-\mathcal{\bm A}||^2_2 \right\}.
\end{equation}

Then the regularized ReLU can be given by a nonnegative constraint ROF model:
\begin{equation}\label{eq:RRELU}
\underset{\mathcal{\bm A} \geqslant 0}{\arg\min}\left\{\frac{1}{2}|| \bm o-\mathcal{\bm A}||^2_2 +\lambda \text{TV}(\mathcal{\bm A})\right\}.
\end{equation}

Similar to the regularized softmax, the problem in \Cref{eq:RRELU} can be solved as follows:
\begin{equation}
\begin{aligned}
(\mathcal{\bm A}^*, \bm \eta^*)=&\arg\underset{\mathcal{\bm A} \geqslant \bm 0 }{\min}~~\underset{ \bm \xi\in\mathbb{B}}{\max}
\{\frac{1}{2}|| \bm o-\mathcal{\bm A}||^2_2
+\lambda <\mathcal{\bm A}, div \bm \xi>\}. \\
\end{aligned}
\end{equation}

The similar primal-dual gradient algorithm could be used to find the solution in an iterative way: 
\begin{equation}\label{eq:rrelu}
\left\{
\begin{array}{ll}
\bm \xi^{t+1} &=\bm  \xi^t - \tau\lambda\triangledown \mathcal{\bm A}^{t} , \\
\bm \eta^{t+1} &= \mathcal{P}_\mathbb{B}(\bm \xi^{t+1}), \\
\mathcal{\bm A}^{t+1} &= \max(\bm 0, \bm o-\lambda div(\bm \eta^{t+1})), \\
\end{array}
\right.
\end{equation}

Once we get convergent $\mathcal{\bm A}$ and $\bm \eta$, the new regularized ReLU function has the following simple  expression:
\begin{equation}
\mathcal{\bm A}=\text{ReLU}(\bm o-\lambda div \bm \eta).
\end{equation}

Usually, dozens of ReLU layers are employed to process feature maps from low level to high level in a CNN. The computational burden will be extremely high if we compute convergent $\mathcal{\bm A}$ and $\bm \eta$ for every active layer. In this paper, we just consider the activation function in the last layer as to be the TV regularized softmax function in image segmentation problem.

\subsection{Backpropagation of Regularized Softmax}
Given initial $\mathcal{\bm A}^{0}$ and $\bm \xi^0$ in the forward propagation stage, we propagate $\bm o$ through \Cref{eq:dualalgo} to achieve a regularized $\bm o$. When doing backpropagation, we need to compute the gradient of loss $\mathcal{L}$ with respect to $\bm o$. Since \Cref{eq:dualalgo} is computed $t+1$ iterations in the forward propagation, we compute the gradients in an inverse order.

When $k=1,\ldots,t+1$, $\bm \eta^{k}$ is the input to compute $\mathcal{\bm A}^{k}$, so we have
\begin{equation}\label{eq:L_eta}
\begin{array}{ll}
\frac{\partial \mathcal{L}}{\partial \bm \eta^{k}}
=& \frac{\partial \mathcal{L}}{\partial \mathcal{\bm A}^{k}}\cdot \frac{\partial \mathcal{\bm A}^{k}}{\partial \bm \eta^{k}}, \ k=1,\ldots,t+1.
\end{array}
\end{equation}

Similarly, $\bm \xi^{t+1}$ is the input to compute  $\bm \eta^{t+1}$, when $k=1,\ldots,t$,  $\bm \xi^{k}$ is the input to compute  $\bm \eta^{k}$ and $\bm \xi^{k+1}$, so we have
\begin{equation}\label{eq:L_xi}
\frac{\partial \mathcal{L}}{\partial \bm \xi^{k}} =
\left\{
\begin{array}{ll}
&\frac{\partial \mathcal{L}}{\partial \mathcal{\bm \eta}^{k}}\cdot \frac{\partial \mathcal{\bm \eta}^{k}}{\partial \bm \xi^{k}}, \ k=t+1\\
&\frac{\partial \mathcal{L}}{\partial \mathcal{\bm \eta}^{k}}\cdot \frac{\partial \mathcal{\bm \eta}^{k}}{\partial \bm \xi^{k}}+ \frac{\partial \mathcal{L}}{\partial \mathcal{\bm \xi}^{k+1}}, \ k=1,\ldots,t. \\
\end{array}
\right .
\end{equation}

As well, when $k=0,\ldots,t$, $\mathcal{\bm A}^{k}$ is the input to compute $\bm \xi^{k+1}$, so we have
\begin{equation}\label{eq:L_A}
\begin{array}{ll}
\frac{\partial \mathcal{L}}{\partial \mathcal{\bm A}^{k}}
=& \frac{\partial \mathcal{L}}{\partial \bm \xi^{k+1}}\cdot \frac{\partial \bm \xi^{k+1}}{\partial \mathcal{\bm A}^{k}}, \ k=0,\ldots,t.
\end{array}
\end{equation}

When $k=0,\ldots,t+1$, $\bm o$ is the input to compute $\mathcal{\bm A}^k$. Given $\mathcal{\bm A}^0=\mathcal{S}(\bm o)$, we have
\begin{equation}\label{eq:L_o}
\begin{array}{ll}
\frac{\partial \mathcal{L}}{\partial \bm o}
=& \frac{\partial \mathcal{L}}{\partial \mathcal{\bm A}^{0}} \cdot \mathcal{S}^{'}(\bm o) + \sum_{k=1}^{t+1}\frac{\partial \mathcal{L}}{\partial \mathcal{\bm A}^{k}} \cdot \mathcal{S}^{'}(\bm o-\lambda div (\bm  \eta^{k})).
\end{array}
\end{equation}

During the backpropagation stage, $\frac{\partial \mathcal{L}}{\partial \mathcal{\bm A}^{t+1}}$ could be given by the loss layer, so we can iteratively obtain $\frac{\partial \mathcal{L}}{\partial \bm \eta^{t+1}},\frac{\partial \mathcal{L}}{\partial \bm \xi^{t+1}},\frac{\partial \mathcal{L}}{\partial \mathcal{\bm A}^{t}},\ldots,\frac{\partial \mathcal{L}}{\partial \bm \eta^{1}}, \frac{\partial \mathcal{L}}{\partial \bm \xi^{1}}, \frac{\partial \mathcal{L}}{\partial \mathcal{\bm A}^{0}}$ by \Cref{eq:L_eta}, \Cref{eq:L_xi} and \Cref{eq:L_A}.

Finally, we can get $\frac{\partial \mathcal{L}}{\partial \bm o}$ by \Cref{eq:L_o}.

\subsection{Implementation Details}
During training stage, we have to compute \Cref{eq:dualalgo} tens to hundreds times for each CNN iteration in order to obtain convergent $\mathcal{\bm A}$ and $\bm \eta$ in the forward propagation stage. Usually, CNN needs dozens of thousands iterations to converge. It means that \Cref{eq:dualalgo} will be computed million times during the whole training stage, which is a huge computation burden. What's more,  we have to iteratively compute $\frac{\partial \mathcal{L}}{\partial \bm \eta^{t+1}},\frac{\partial \mathcal{L}}{\partial \bm \xi^{t+1}},\frac{\partial \mathcal{L}}{\partial \mathcal{\bm A}^{t}},\ldots,\frac{\partial \mathcal{L}}{\partial \bm \eta^{1}}, \frac{\partial \mathcal{L}}{\partial \bm \xi^{1}}, \frac{\partial \mathcal{L}}{\partial \mathcal{\bm A}^{0}}$ and keep those matrices in memory in backpropagation stage. Numerous computation and memory resources are required. The period for training one batch of models will be as long as weeks. And the mini-batch size will be smaller due to more memory is required for training each image. However, smaller mini-batch size may lead to decline in accuracy.

Currently, we compute \Cref{eq:dualalgo} just once for each training iteration and it will be fully performed during the testing stage. This is a trade-off between regularization effect and computation, memory resources. Though there will be less regularization effect, the demand for computation and memory resources during training stage is greatly reduced. Visible regularization effect is still observed in out experimental results in section \Cref{sec:results}.

In order to keep consistent with the one iteration \Cref{eq:dualalgo} in the forward propagation, we design a step-by-step strategy to compute $\frac{\partial \mathcal{L}}{\partial{\bm o}}$ in the backward propagation stage.  In all computation, we perform \Cref{eq:dualalgo} just one iteration and we'll get
\begin{equation}\label{eq:TV}
\left\{
\begin{array}{ll}
\bm \xi &= \bm \xi_0 - \tau\lambda\triangledown\mathcal{S}(\bm o-\lambda div(\bm \eta_0)) , \\
\bm \eta &= \mathcal{P}_\mathbb{B}(\bm \xi), \\
\mathcal{\bm A} &= \mathcal{S}(\bm o-\lambda div(\bm \eta)). \\
\end{array}
\right.
\end{equation}

We set the initialization $\bm \xi_0$ and $\bm \eta_0$ to $\bm 0$, respectively. Then the one iteration scheme \Cref{eq:TV} could be simplified as :
\begin{equation}\label{eq:TV1}
\left\{
\begin{array}{ll}
\bm \xi &= -\tau\lambda\triangledown\mathcal{S}(\bm o) , \\
\bm \eta &= \mathcal{P}_\mathbb{B}(\bm \xi), \\
\mathcal{\bm A} &= \mathcal{S}(\bm o-\lambda div(\bm \eta))
\end{array}
\right.
\end{equation}

Let $\mathcal{L}$ be the loss function, according to the backpropagation scheme \Cref{bp}, the gradient of $\mathcal{L}$ with respect to $\bm o$ in \Cref{eq:TV1} is computed as:
\begin{equation}\label{eq:gradient_oK}
\begin{array}{rl}
\frac{\partial \mathcal{L}}{\partial{\bm o}} &=\frac{\partial \mathcal{L}}{\partial \mathcal{\bm A}}\cdot \frac{\partial \mathcal{\bm A}}{\partial {\bm o}},\\
&= \frac{\partial \mathcal{L}}{\partial \mathcal{\bm A}}\cdot(\mathcal{S}^{'}(\bm o - \lambda div(\bm \eta))+\frac{\partial \mathcal{\bm A}}{\partial {\bm \eta}} \cdot \frac{\partial {\bm \eta}}{\partial {\bm o}}),\\
&=\frac{\partial \mathcal{L}}{\partial \mathcal{\bm A}}\cdot \mathcal{S}^{'}(\bm o - \lambda div(\bm \eta)) +\frac{\partial \mathcal{L}}{\partial {\bm \eta}}\cdot\frac{\partial {\bm \eta}}{\partial {\bm o}}, \\
\text{where   } \ \frac{\partial {\bm \eta}}{\partial {\bm o}} &=-\tau\lambda \frac{\partial \triangledown\mathcal{S}}{\partial \mathcal{S}}\cdot \frac{\partial \mathcal{P}_\mathbb{B}}{\partial {\bm \xi}}. \\
\end{array}
\end{equation}

In \Cref{eq:gradient_oK} we can see that $\bm \eta$ contributes to updating gradients during the backpropagation stage. This is quite different from other post-processing methods such as CRF.

$\frac{\partial \mathcal{L}}{\partial{\bm o}^{k+1}}$ in \Cref{bp} will be updated by \Cref{eq:gradient_oK} for the proposed regularized network.
Since the item $\tau\lambda$ in \Cref{eq:TV1} and \Cref{eq:gradient_oK} could be seen as a scaled step size. In our implementations, we define it as a new constant and fix it by manual tuning.  The regularization parameter $\lambda$ will be learned as explained in the next section. 

\subsection{Training of the Regularization Parameter $\lambda$}
The regularization parameter $\lambda$ controls regularization effect. When it is too large, the output may be over-regularized, leading to a drop in accuracy. When it is too small, the TV item will contribute little to training. Generally, we manually set different $\lambda$ and select a best one. However, it could be quite boring and inefficient to try different $\lambda$ for each CNN on each dataset. Here, we introduce a training scheme to select $\lambda$ automatically instead of manual setting. This will also help improve the training procedure.

The gradient of $\mathcal{L}$ with respect to $\lambda$ is in the following:
\begin{equation}\label{eq:gradient_lambda}
\begin{aligned}
\frac{\partial \mathcal{L}}{\partial \lambda} &= \frac{\partial \mathcal{\bm A}}{\partial\lambda} \cdot\frac{\partial \mathcal{L}}{\partial \mathcal{\bm A}} = -div(\bm \eta)\cdot\mathcal{S}^{'}(\bm o-\lambda div(\bm \eta))\cdot\frac{\partial \mathcal{L}}{\partial \mathcal{\bm A}}.
\end{aligned}
\end{equation}

When doing backpropagation in each iteration during training stage, we both update the parameter set $\Theta$ and $\lambda$ by gradient descend method simultaneously. $\lambda$ is updated as follows:
\begin{equation}\label{eq:update_lambda}
\lambda^{step+1} = \lambda^{step} - \tau_\lambda\frac{\partial \mathcal{L}}{\partial \lambda}\bigg|_{\lambda=\lambda^{step}},
\end{equation}

where $\tau_\lambda$ is the learning rate for $\lambda$, $step$ is the training iteration number.

\FloatBarrier
\section{Experimental Results}
\label{sec:results}

We quantify the performance of regularized softmax on Unet and Segnet using Caffe implementation. Since Unet is prominent in biomedical image segmentation, Unet and Unet with regularized softmax activation function (RUnet) are tested on White Blood Cell Dataset. Segnet and Segnet with regularized softmax activtion function (RSegnet) are tested on CamVid Dataset and SUN-RGBD Dataset.

We use SGD solver with momentum of 0.9 for each network. The learning rates of Unet and RUnet are fixed to be 0.0001, their weights are both randomly initialized. The learning rates of Segnet and RSegnet are fixed to be 0.001, their weights are both initialized from the VGG model trained on ImageNet using the techniques described in He et al.\cite{he2015delving}, the same as the author of Segnet did. 

During testing stage, we first train those four networks on clean training dataset and test them on both clean and noisy images in order to further evaluate the robustness of our proposed method when encountering noise. Adding noise to training images is a common technique to make networks robust to noise, we also train the four networks on noisy data to make further comparison. We choose global accuracy and mean intersection over union (mIoU) to be our quantitative measures.

Given a segmentation result $u$, we evaluate its regularization effect as follows:
\begin{equation}
\begin{aligned}
RE(u) = \frac{100}{N_1\times N_2} \sum_{i=1}^{N_1}\sum_{j=1}^{N_2}|\triangledown u_{i,j}|
\end{aligned}
\end{equation}
where $N_1,N_2$ are the width and height of $u$, respectively. $\triangledown u$ is defined as follows:
\begin{equation}
\begin{aligned}
\label{eq:gradient}
(\triangledown u)_{i,j} = ((\triangledown u)^1_{i,j},(\triangledown u)^2_{i,j}).
\end{aligned}
\end{equation}

Segmentation results with higher RE means lower regularization effect, which often have more isolated small regions and serrated edges.  

\subsection{WBC Dataset}
White Blood Cell Image Dataset\cite{Zheng2018} consists of two sub-datasets. Dataset 1 contains three hundred 120x120 images and their color depth is 24 bits. Dataset 2 consists of one hundred 300x300 color images. The cell images are generally purple and may contain many red blood cells around the white blood cells. Since the image size of Dataset 1 is a little bit small, it is not suitable for deep CNNs like Unet. We select Dataset 2 as our experimental dataset.

\begin{figure}[tbhp]
	\centering
	\includegraphics[width=0.8\linewidth]{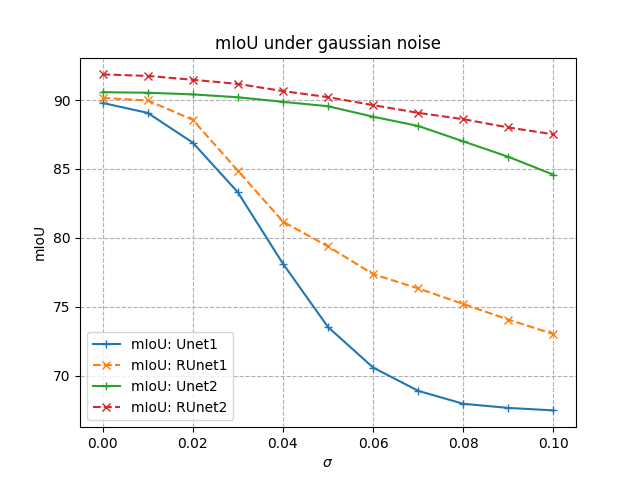}
	\caption{Unet1 and RUnet1 are trained on clean WBC dataset, Unet2 and RUnet2 are trained on noisy WBC dataset. We add gaussian noise with zero mean, standard deviation $\sigma$ from 0.01 to 0.1 to WBC testing dataset.} 
	\label{fig:3}
\end{figure}

WBC Dataset 2 has simple image structure and distinct details, it is very convenient for us to observe the difference in details intuitively. We replace original softmax layer with regularized softmax layer, other layers and parameters of Unet remain the same.

\setlength{\tabcolsep}{1.2mm}{
	\renewcommand\arraystretch{1.2}
	
	\begin{table}[tbhp]
		\caption{Results of Unet1 and RUnet1 trained on clean dataset.}\label{table1}
		\begin{center}
			\begin{tabular}{cccccccccc}
				\hline
				& Method & \multicolumn{1}{c}{clean} & \multicolumn{5}{c}{gaussain} & \multicolumn{1}{c}{pepper} & \multicolumn{1}{c}{salt} \\
				\hline
				noise level	 &	&	& 0.01 & 0.03 & 0.05 & 0.07 & 0.09 & 0.01 & 0.01   \\		
				\hline
				\multirow{2}{*}{mIoU}     & Unet1 & 89.79 & 89.02 & 83.19 & 73.94 & 68.74 & 67.67 & 80.68 & 77.75   \\			
				& RUnet1 & \bf 90.15 & \bf 90.01 & \bf 84.95 & \bf 79.25 & \bf 76.14 & \bf 74.20 & \bf 85.46 & \bf 84.07  \\			
				\hline
				\multirow{2}{*}{Accuracy} & Unet1 & 97.04 & 96.69  & 93.55 & 86.43 & 81.22  & 80.33 & 92.55 & 90.00   \\			
				& RUnet1 & \bf 97.13 & \bf 97.04 & \bf 94.35 & \bf 90.54 & \bf 88.06  & \bf 86.55 & \bf 94.87 & \bf 96.26  \\			
				\hline
				\multirow{2}{*}{RE}       & Unet1 & 1.82 & 1.94 & 3.50 & 7.80 & 11.07 & 13.60 & 5.99 & 4.39   \\			
				& RUnet1 & \bf 1.30 & \bf 1.30 & \bf 1.39 & \bf 1.48 & \bf 1.52 & \bf 1.56 & \bf 1.35 & \bf 1.32  \\			
				\hline
			\end{tabular}
		\end{center}
	\end{table}

	\begin{table}[tbhp]
	\caption{Results of Unet2 and RUnet2 trained on noisy dataset.}\label{table2}
	\begin{center}
		\begin{tabular}{cccccccccc}
			\hline
			& Method & \multicolumn{1}{c}{clean} & \multicolumn{5}{c}{gaussain} & \multicolumn{1}{c}{pepper} & \multicolumn{1}{c}{salt} \\
			\hline
			noise level	 &	&	& 0.01 & 0.03 & 0.05 & 0.07 & 0.09 & 0.01 & 0.01   \\
			\hline
			\multirow{2}{*}{mIoU}     & Unet2 & 90.57 & 90.52 & 90.31 & 89.49 & 88.07 & 85.87 & 88.20 & 88.84   \\
			& RUnet2 & \bf 91.86 & \bf 91.73 & \bf 91.19 & \bf 90.24 & \bf 89.29 & \bf 88.37 & \bf 89.85 & \bf 91.22  \\		
			\hline
			\multirow{2}{*}{Accuracy} & Unet2 & 97.25 & 97.23 & 97.14 & 96.89 & 96.47 & 95.77 & 96.54 & 96.77   \\	
			& RUnet2 & \bf 97.64 & \bf 97.60 & \bf 97.38 & \bf 97.00 & \bf 96.67 & \bf 96.38 & \bf 96.96 & \bf 97.37  \\				
			\hline
			\multirow{2}{*}{RE}       & Unet2 & 1.79 & 1.80 & 1.87 & 2.02 & 2.17 & 2.87 & 2.38 & 2.19   \\
			& RUnet2 & \bf 1.32 & \bf 1.32 & \bf 1.33  & \bf 1.35 & \bf 1.35 & \bf 1.34 & \bf 1.32 & \bf 1.34  \\			
			\hline
		\end{tabular}
	\end{center}
\end{table}
	
	\begin{figure*}[tbhp]
		\centering
		\begin{subfigure}
			{
				\begin{minipage}[b]{0.075\textwidth}
					noisy image  \ \vspace{1.2\textwidth} \\
					ground truth \ \vspace{1.2\textwidth} \\
					Unet		 \ \vspace{1.0\textwidth} \\
					Unet \& post-TV      \ \vspace{0.8\textwidth} \\
					RUnet 		 \ \vspace{0.55\textwidth}
				\end{minipage}
			}
		\end{subfigure}
		\begin{subfigure}
			{
				\begin{minipage}[b]{0.145\textwidth}
					\includegraphics[width=1\textwidth]{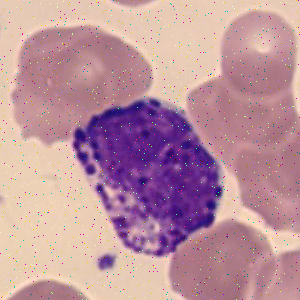} \\
					\includegraphics[width=1\textwidth]{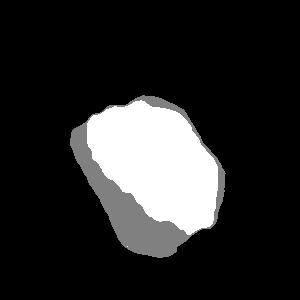} \\
					\includegraphics[width=1\textwidth]{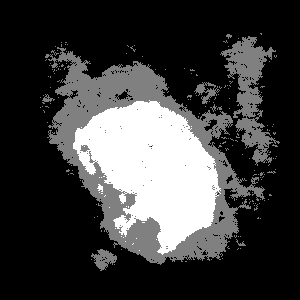} \\
					\includegraphics[width=1\textwidth]{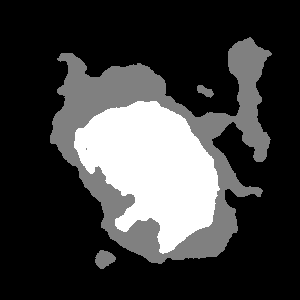} \\
					\includegraphics[width=1\textwidth]{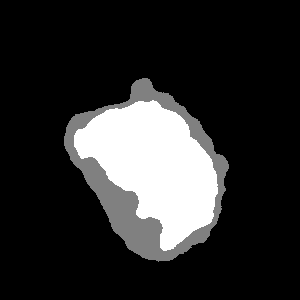}
				\end{minipage}
			}
		\end{subfigure}
		\begin{subfigure}
			{
				\begin{minipage}[b]{0.145\textwidth}
					\includegraphics[width=1\textwidth]{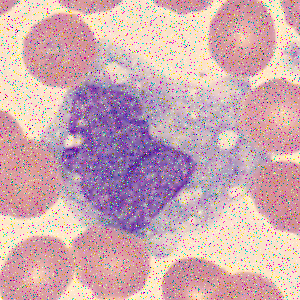} \\
					\includegraphics[width=1\textwidth]{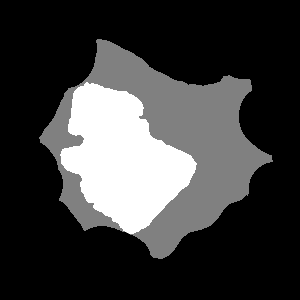} \\
					\includegraphics[width=1\textwidth]{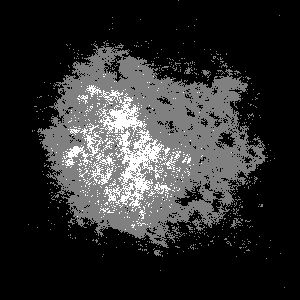} \\
					\includegraphics[width=1\textwidth]{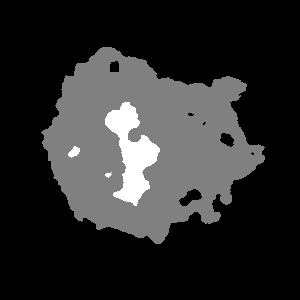} \\
					\includegraphics[width=1\textwidth]{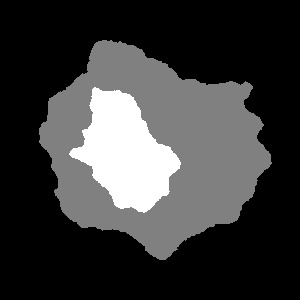}
				\end{minipage}
			}
		\end{subfigure}
		\begin{subfigure}
			{
				\begin{minipage}[b]{0.145\textwidth}
					\includegraphics[width=1\textwidth]{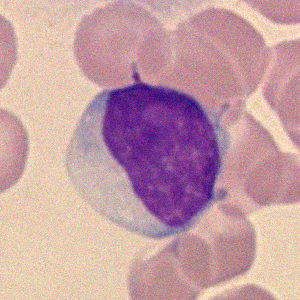} \\
					\includegraphics[width=1\textwidth]{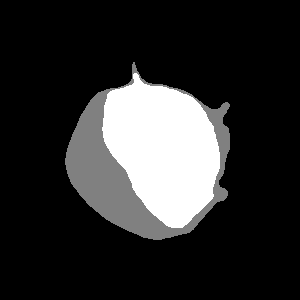} \\
					\includegraphics[width=1\textwidth]{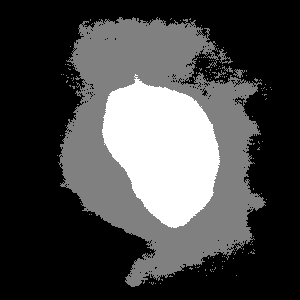} \\
					\includegraphics[width=1\textwidth]{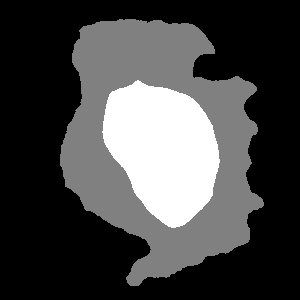} \\
					\includegraphics[width=1\textwidth]{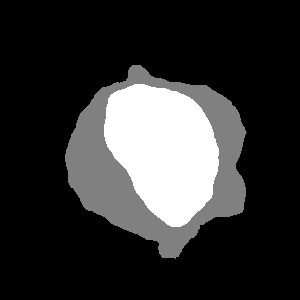}
				\end{minipage}
			}
		\end{subfigure}
		\begin{subfigure}
			{
				\begin{minipage}[b]{0.145\textwidth}
					\includegraphics[width=1\textwidth]{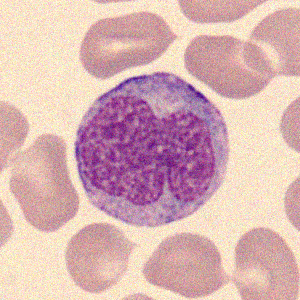} \\
					\includegraphics[width=1\textwidth]{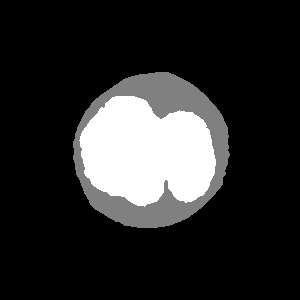} \\
					\includegraphics[width=1\textwidth]{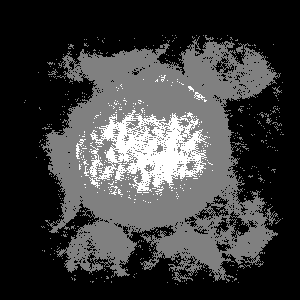} \\
					\includegraphics[width=1\textwidth]{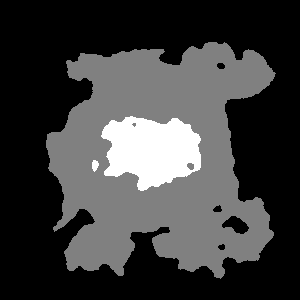} \\
					\includegraphics[width=1\textwidth]{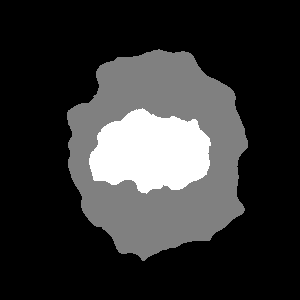}
				\end{minipage}
			}
		\end{subfigure}
		\begin{subfigure}
			{
				\begin{minipage}[b]{0.145\textwidth}
					\includegraphics[width=1\textwidth]{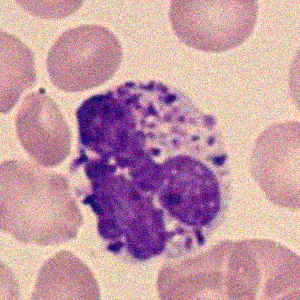} \\
					\includegraphics[width=1\textwidth]{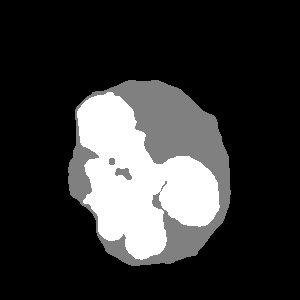} \\
					\includegraphics[width=1\textwidth]{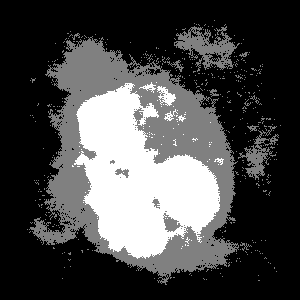} \\
					\includegraphics[width=1\textwidth]{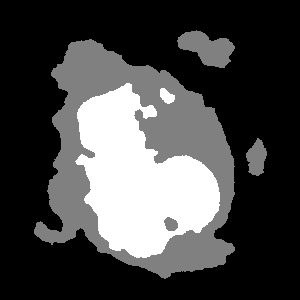} \\
					\includegraphics[width=1\textwidth]{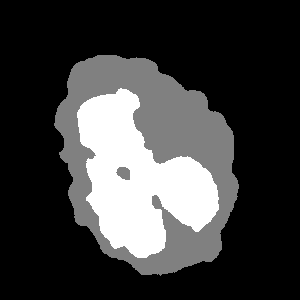}
				\end{minipage}
			}
		\end{subfigure}
		\caption{Segmentation results predicted by Unet and RUnet trained on noisy dataset. Noise type from left to right: small level salt and pepper(s\&p) noise, large level s\&p noise, small level gaussian noise, medium level gaussian noise, medium level gaussian noise.}
		\label{fig:segmentation_task_2}
	\end{figure*}
	
	We randomly pick out 60 images from Dataset 2 as training dataset, the other 40 images are used for testing. Both Unet and RUnet are trained for 20k iterations, their mini-batch sizes are both eight.

	As we cannot always obtain very clean images in practice, we want to know  how the segmentation result will change when encountering small noise. First, Unet, RUnet, Segnet and RSegnet are all trained on clean data and tested on both clean and noise data with different noise. In \Cref{fig:3} we can see that mIoU of Unet1 has a significant drop when the noise level increases. However, the degradation in mIoU of RUnet1 is greatly alleviated. The benefit in mIoU from regularized softmax layer is very impressive. The mIoU curve of Unet1 seems to be convergent when $\sigma$ is greater than $0.8$ because of the majority pixels are recognized as background.
	
	When the training dataset contains images with noise, the trained model can be more robust. We randomly pick out 20 images in training dataset and randomly add gaussian noise with zero mean, $\sigma=0.05$ or pepper and salt noise with 1\% pixels' value changed to each image. We make a further comparison when Unet and RUnet are trained on noisy dataset. \Cref{table1} and \Cref{table2} show predictions on clean data and data with different noise levels: gaussian noise with zero mean, standard deviations $\sigma=0.01,0.03,0.05,0.07,0.09$, pepper and salt noise with 1\% pixels' value changed per image.
	
	In \Cref{fig:3} we can see that the loss of performance is greatly reduced when adding some noise to training images. In \Cref{table2}, performance of RUnet model is still better than Unet model. We also apply post-TV processing to segmentation results of Unet, that is we replace softmax with regularized softmax during testing stage and perform \Cref{eq:dualalgo} 100 iterations for each prediction. In \Cref{fig:segmentation_task_2} row 4, we can see that the segmentation results have little improvement after post-TV processing, and it is still inferior to RUnet. This shows that our proposed regularized softmax help Unet find a better local optimum. Although post-TV processing could also bring regularization effect, it doesn't contribute to updating the model weights during training stage and its $\lambda$ is manually set and thus not learnable, over-regularization may happen. Our trainable $\lambda$ scheme helps avoid falling into such a problem.
	
	We can see obvious degradation in predictions on noisy images from row 3 in \Cref{fig:segmentation_task_2}. However, regularized softmax alleviates this problem, the segmentation results of RUnet are much better. We also observe that edges in segmentation results provided by RUnet are very smooth and there are no isolated points.
	
	We have tried to select a best $\lambda$ for post-TV processing and there are much less isolated points and regions in \Cref{fig:segmentation_task_2} row 4. But over-regularization also happens due to manually set $\lambda$. In \Cref{fig:segmentation_task_2} column 5, we can see that there are small holes inside the nucleus in ground truth. However, after applying post-TV processing, those holes just disappears in segmentation result at row 4 due to over-regularization. The segmentation result at row 5 preserves this detail.

	\FloatBarrier
	\subsection{CamVid Dataset}
	CamVid Dataset\cite{johnson2016driving} is a collection of videos with object class semantic labels. This sequence depicts a moving driving scene in the city of Cambridge filmed from a moving car. It is a challenging dataset and selected as the benchmark dataset of Segnet. The authors of Segnet use an 11 class version with an image size of 360 by 480, and they pick out 367 frames as training images and 233 frames as testing images.
	
	\begin{figure}[tbhp]
		\centering
		\includegraphics[width=0.8\linewidth]{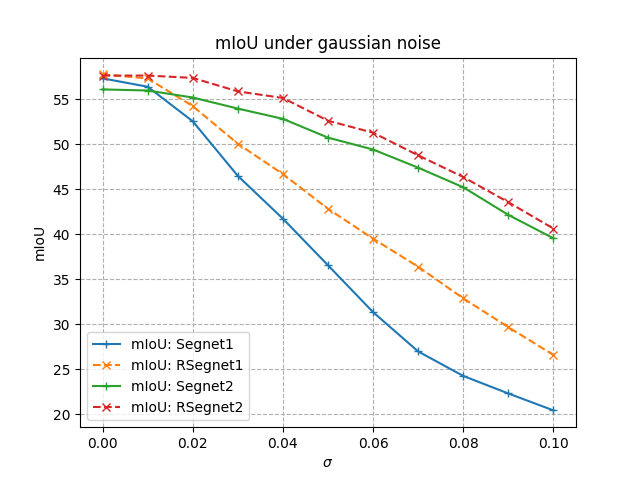}
		\caption{Segnet1 and RSegnet1 are trained on clean CamVid dataset, Segnet2 and RSegnet2 are trained on noisy CamVid dataset. We add gaussian noise with zero mean, standard deviation $\sigma$ from 0.01 to 0.1 to CamVid testing dataset.}
		\label{fig:5}
	\end{figure}
	
	We replace original softmax layer with regularized softmax layer, other layers and parameters of Segnet and RSegnet remain the same. Both Segnet and RSegnet are trained for 80k iterations with weights initialized from the VGG model trained on ImageNet, their mini-batch sizes are both four. Learning rates of Segnet and RSegnet are fixed to be 0.001.
	
	\begin{table}[tbhp]
		\caption{Results of Segnet1 and RSegnet1 trained on clean dataset.}\label{table3}
		\begin{center}
			\begin{tabular}{cccccccccc}
				\hline
				& Method & \multicolumn{1}{c}{clean} & \multicolumn{5}{c}{gaussain} & \multicolumn{1}{c}{pepper} & \multicolumn{1}{c}{salt} \\
				\hline
				noise level	 &	&	& 0.01 & 0.03 & 0.05 & 0.07 & 0.09 & 0.01 & 0.01   \\		
				\hline
				\multirow{2}{*}{mIoU}     & Segnet1 & 57.35 & 56.42 & 46.51 & 36.57 & 26.99 & 22.32 & 50.42 & 35.45   \\			
				& RSegnet1 & \bf 57.79 & \bf 57.34 & \bf 50.08 & \bf 42.85 & \bf 36.41 & \bf 29.69 & \bf 52.80 & \bf 41.16 \\
				\hline
				\multirow{2}{*}{Accuracy} & Segnet1 & 87.74 & 87.42  & 81.71 & 71.27 & 55.82  & 48.92 & 81.52 & 71.33   \\			
				& RSegnet1 & \bf 88.01 & \bf 87.74 & \bf 84.18 & \bf 78.44 & \bf 72.20  & \bf 63.12 & \bf 84.17 & \bf 78.85  \\			
				\hline
				\multirow{2}{*}{RE}       & Segnet1 & 4.10 & 4.18 & 5.62 & 7.79 & 9.00 & 9.63 & 5.88 & 7.04   \\			
				& RSegnet1 & \bf 2.43 & \bf 2.44 & \bf 2.69 & \bf 3.20 & \bf 3.78 & \bf 4.27 & \bf 2.69 & \bf 3.11  \\			
				\hline
			\end{tabular}
		\end{center}
	\end{table}
	
	\begin{table}[tbhp]
		\caption{Results of Segnet2 and RSegnet2 trained on noisy dataset.}\label{table4}
		\begin{center}
			\begin{tabular}{cccccccccc}
				\hline
				& Method & \multicolumn{1}{c}{clean} & \multicolumn{5}{c}{gaussain} & \multicolumn{1}{c}{pepper} & \multicolumn{1}{c}{salt} \\
				\hline
				noise level	 &	&	& 0.01 & 0.03 & 0.05 & 0.07 & 0.09 & 0.01 & 0.01   \\
				\hline
				\multirow{2}{*}{mIoU}     & Segnet2 & 56.12 & 56.00 & 53.99 & 50.76 & 47.42 & 42.17 & 55.54 & 50.82   \\
				& RSegnet2 & \bf 57.66 & \bf 57.66 & \bf 55.90 & \bf 52.64 & \bf 48.79 & \bf 43.58 & \bf 57.21 & \bf 51.81  \\		
				\hline
				\multirow{2}{*}{Accuracy} & Segnet2 & 87.54 & 87.52 & 86.95 & 85.56 & 83.43 & 79.27 & 87.22 & 85.90   \\	
				& RSegnet2 & \bf 88.10 & \bf 88.09 & \bf 87.63 & \bf 86.12 & \bf 83.56 & \bf 79.36 & \bf 87.68 & \bf 86.05  \\				
				\hline
				\multirow{2}{*}{RE}       & Segnet2 & 4.56 & 4.52 & 4.57 & 4.82 & 5.41 & 6.60 & 4.51 & 4.71   \\
				& RSegnet2 & \bf 2.51 & \bf 2.49 & \bf 2.49  & \bf 2.59 & \bf 2.93 & \bf 3.39 & \bf 2.46 & \bf 2.50  \\			
				\hline
			\end{tabular}
		\end{center}
	\end{table}
	
	First, Segnet and RSegnet are all trained on clean data and tested on data with different noise.  Similar degradation could be found in \Cref{fig:5}. In \Cref{fig:segmentation_task_4} we can see the prediction results of bicyclist and road become very messy in segmentation results predicted by Segnet at column 1 and column 4, whereas RSegnet provides relative good segmentation results.  
	
		\begin{figure*}[tbhp]
		\centering
		\begin{subfigure}
			{
				\begin{minipage}[b]{0.09\textwidth}
					test image  \ \vspace{1.0\textwidth} \\
					ground truth \ \vspace{1.0\textwidth} \\
					Segnet		 \ \vspace{1.0\textwidth} \\
					Segnet with post-TV      \ \vspace{0.5\textwidth} \\
					RSegnet 	 \ \vspace{0.2\textwidth} \\
				\end{minipage}
			}
		\end{subfigure}
		\begin{subfigure}
			{
				\begin{minipage}[b]{0.19\textwidth}
					\includegraphics[width=1\textwidth]{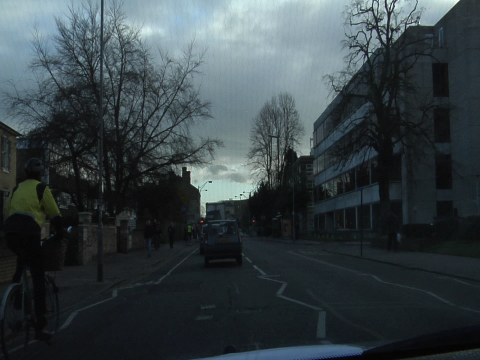} \\
					\includegraphics[width=1\textwidth]{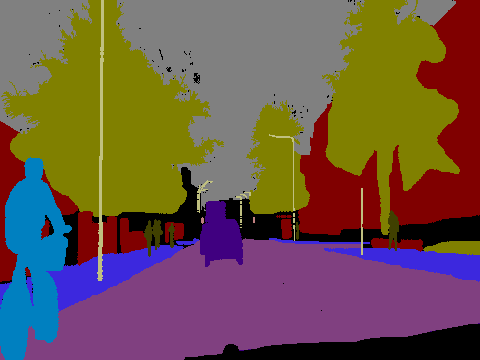} \\
					\includegraphics[width=1\textwidth]{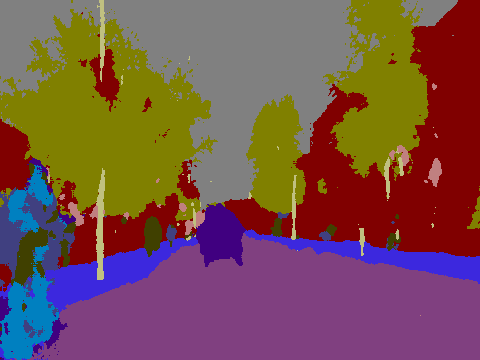} \\
					\includegraphics[width=1\textwidth]{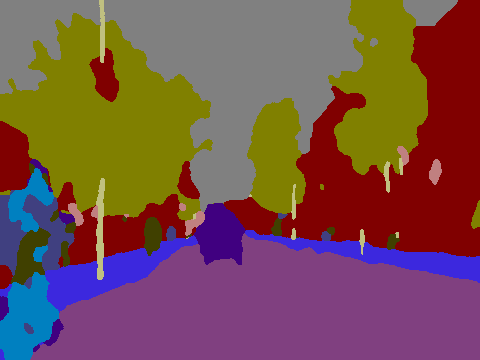} \\
					\includegraphics[width=1\textwidth]{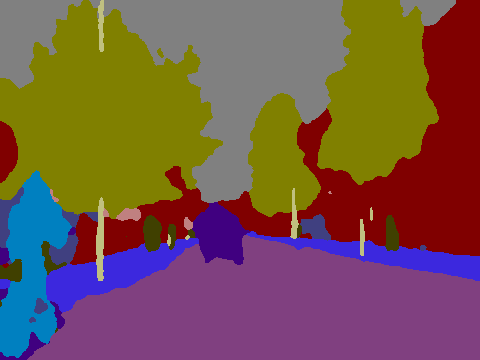}
				\end{minipage}
			}
		\end{subfigure}
		\begin{subfigure}
			{
				\begin{minipage}[b]{0.19\textwidth}
					\includegraphics[width=1\textwidth]{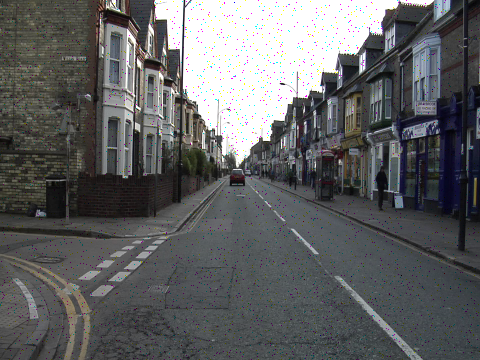} \\
					\includegraphics[width=1\textwidth]{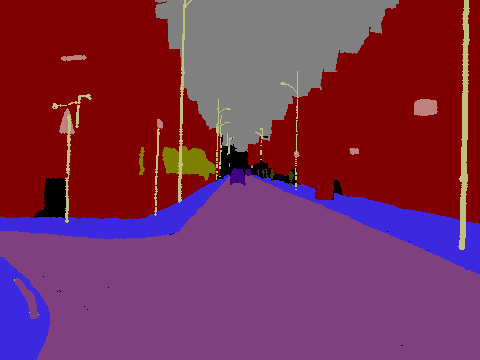} \\
					\includegraphics[width=1\textwidth]{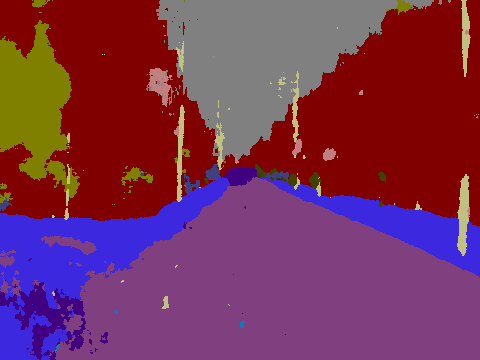} \\
					\includegraphics[width=1\textwidth]{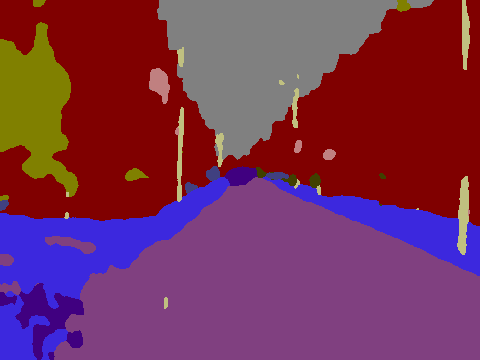} \\
					\includegraphics[width=1\textwidth]{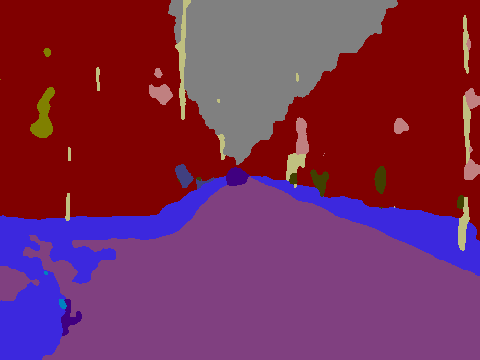}
				\end{minipage}
			}
		\end{subfigure}
		\begin{subfigure}
			{
				\begin{minipage}[b]{0.19\textwidth}
					\includegraphics[width=1\textwidth]{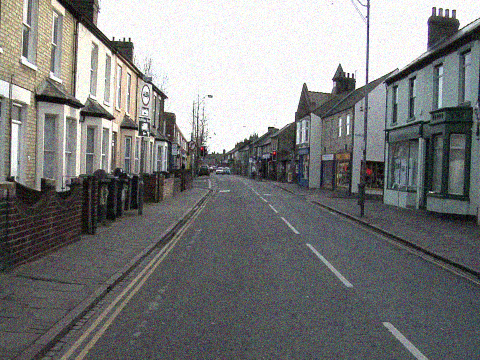} \\
					\includegraphics[width=1\textwidth]{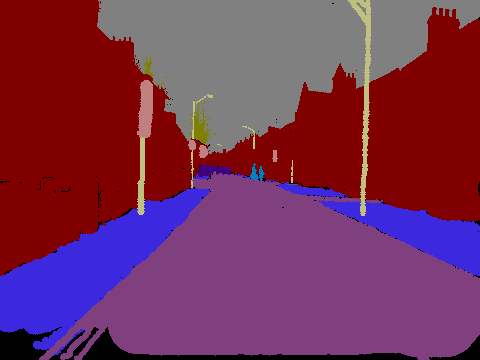} \\
					\includegraphics[width=1\textwidth]{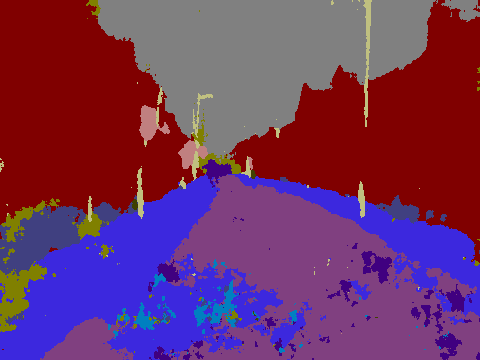} \\
					\includegraphics[width=1\textwidth]{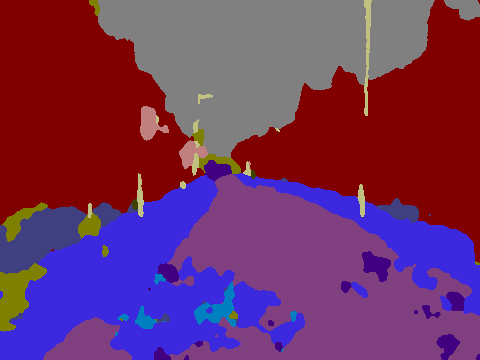} \\
					\includegraphics[width=1\textwidth]{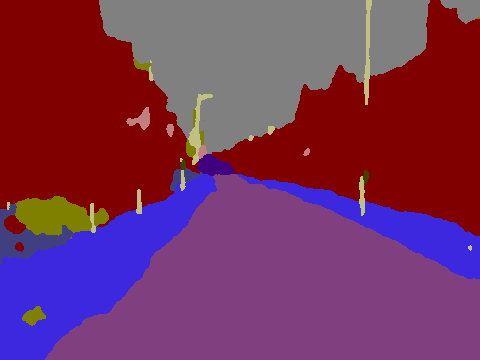}
				\end{minipage}
			}
		\end{subfigure}
		\begin{subfigure}
			{
				\begin{minipage}[b]{0.19\textwidth}
					\includegraphics[width=1\textwidth]{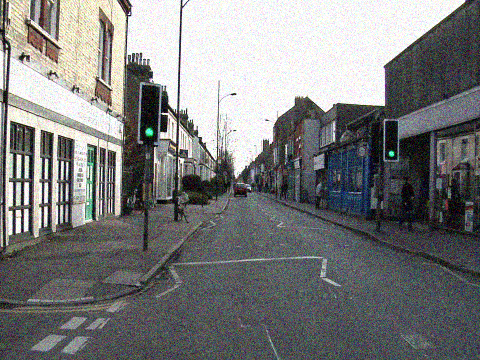} \\
					\includegraphics[width=1\textwidth]{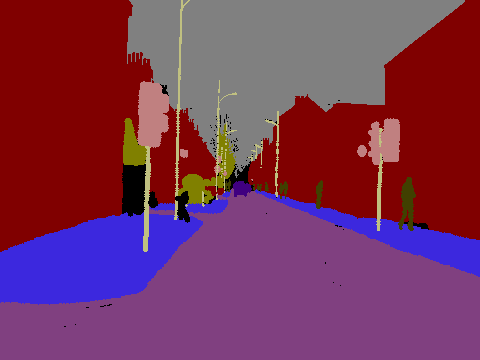} \\
					\includegraphics[width=1\textwidth]{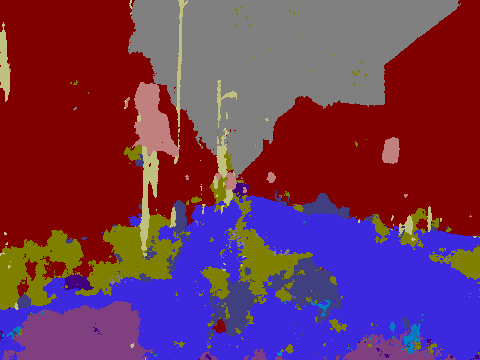} \\
					\includegraphics[width=1\textwidth]{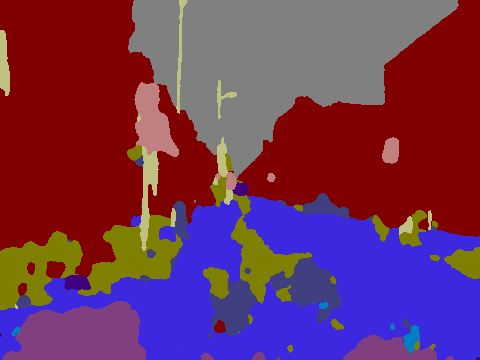} \\
					\includegraphics[width=1\textwidth]{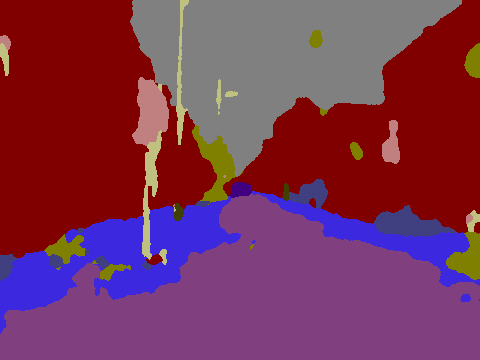}
				\end{minipage}
			}
		\end{subfigure}
		\caption{Segmentation results of Segnet and RSegnet trained on noisy dataset. Noise type from left to right: clean image, medium level pepper noise, medium level gaussian noise, large level gaussian noise.}
		\label{fig:segmentation_task_4}
	\end{figure*}
	
	Then, we randomly pick out 90 images in training dataset and randomly add
	gaussian noise with zero mean, $\sigma=0.03$ or pepper and salt noise with 1\% pixels' value changed to each image. We make a further comparison when Segnet and RSegnet are trained on noisy dataset. \Cref{table3} and \Cref{table4} show predictions on data with different noise levels.
	In  \Cref{fig:5}, we can find that the performance of Segnet and RSegnet on clean testing dataset has a little drop when trained on noisy data, this is due to the different data distribution between training data and testing data. Road scene has very complex image structure, adding noise to training dataset brings further complexity to training task. However, the accuracy, mIoU of Segnet model and RSegnet model appear much more robust to noise. This is a trade-off between accuracy and robustness.
	
	The performance of RSegnet is better than Segnet. Although post-TV processing brings regularization effect to the segmentation results of Segnet, over-regularization also happens. In \Cref{fig:segmentation_task_4} column 3, we can see that the distant column-pole is over-regularized after post-TV processing. Nevertheless, it is well preserved by RSegnet.
	
	\FloatBarrier
	\subsection{SUN-RGBD Dataset}
	SUN-RGBD Dataset\cite{song2015sun} is a much more challenging dataset of
	indoor scenes with 10355 images in total. We randomly select 5,285 images as our training dataset and the remaining images are used as testing dataset. The annotation files contain thousands of labels and we select 37 main categories as our segmentation classes, like the authors of Segnet did. Different sensors are used to capture the scenes and images are in various resolutions. A stochastic patch of size 360 by 480 is cropped from each image.
	
	\begin{figure}[tbhp]
		\centering
		\includegraphics[width=0.8\linewidth]{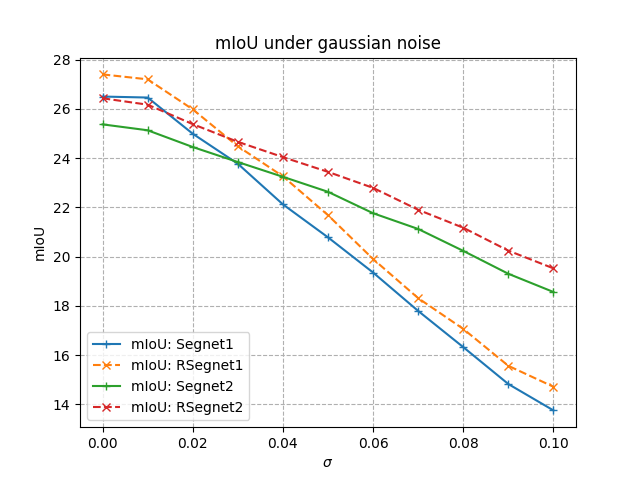}
		\caption{Segnet1 and RSegnet1 are trained on clean SUN-RGBD Dataset, Segnet2 and RSegnet2 are trained on noisy SUN-RGBD dataset. We add gaussian noise with zero mean, standard deviation $\sigma$ from 0.01 to 0.1 to SUN-RGBD testing dataset.}
		\label{fig:6}
	\end{figure}
	
	\begin{table}[tbhp]
		\caption{Results of Segnet1 and RSegnet1 trained on clean dataset.}\label{table5}
		\begin{center}
			\begin{tabular}{cccccccccc}
				\hline
				& Method & \multicolumn{1}{c}{clean} & \multicolumn{5}{c}{gaussain} & \multicolumn{1}{c}{pepper} & \multicolumn{1}{c}{salt} \\
				\hline
				noise level	 &	&	& 0.01 & 0.03 & 0.05 & 0.07 & 0.09 & 0.01 & 0.01   \\		
				\hline
				\multirow{2}{*}{mIoU}     & Segnet1 & 26.50 & 26.46 & 23.77 & 20.78 & 17.79 & 14.82 & 19.79 & 19.28   \\			
				& RSegnet1 & \bf 27.40 & \bf 27.20 & \bf 24.48 & \bf 21.68 & \bf 18.31 & \bf 15.56 & \bf 21.01 & \bf 20.68 \\
				\hline
				\multirow{2}{*}{Accuracy} & Segnet1 & 68.62 & 68.42 & 66.75 & 64.26 & 61.40 & 58.03  & 63.00 & 60.53   \\			
				& RSegnet1 & \bf 68.94 & \bf 68.70 & \bf 66.83 & \bf 64.49 & \bf 61.68  & \bf 58.82 & \bf 63.22 & \bf 60.89  \\			
				\hline
				\multirow{2}{*}{RE}       & Segnet1 & 4.99 & 5.01 & 4.99 & 5.01 & 5.08 & 5.24 & 5.64 & 5.54   \\			
				& RSegnet1 & \bf 2.74 & \bf 2.73 & \bf 2.72 & \bf 2.69 & \bf 2.71 & \bf 2.72 & \bf 2.90 & \bf 2.89  \\			
				\hline
			\end{tabular}
		\end{center}
	\end{table}
	
	\begin{table}[tbhp]
		\caption{Results of Segnet2 and RSegnet2 trained on noisy dataset.}\label{table6}
		\begin{center}
			\begin{tabular}{cccccccccc}
				\hline
				& Method & \multicolumn{1}{c}{clean} & \multicolumn{5}{c}{gaussain} & \multicolumn{1}{c}{pepper} & \multicolumn{1}{c}{salt} \\
				\hline
				noise level	 &	&	& 0.01 & 0.03 & 0.05 & 0.07 & 0.09 & 0.01 & 0.01   \\
				\hline
				\multirow{2}{*}{mIoU}     & Segnet2 & 25.37 & 25.13 & 23.84 & 22.63 & 21.12 & 19.30 & 23.25 & 23.36   \\
				
				& RSegnet2 & \bf 26.43 & \bf 26.17 & \bf 24.66 & \bf 23.44 & \bf 21.90 & \bf 20.24 & \bf 24.37 & \bf 24.42  \\		
				\hline
				\multirow{2}{*}{Accuracy} & Segnet2 & 67.52 & 67.36 & 66.36 & 65.14 & 63.61 & 61.85 & 65.89 & 65.29   \\	
				
				& RSegnet2 & \bf 68.29 & \bf 68.08 & \bf 66.97 & \bf 65.72 & \bf 64.36 & \bf 62.94 & \bf 66.73 & \bf 66.19  \\				
				\hline
				\multirow{2}{*}{RE}       & Segnet2 & 4.84 & 4.84 & 4.91 & 5.03 & 5.21 & 5.40 & 5.03 & 5.10   \\
				
				& RSegnet2 & \bf 2.31 & \bf 2.49 & \bf 2.49  & \bf 2.59 & \bf 2.93 & \bf 3.39 & \bf 2.35 & \bf 2.36  \\			
				\hline
			\end{tabular}
		\end{center}
	\end{table}

	\begin{figure*}[tbhp]
		\centering
		\begin{subfigure}
			{
				\begin{minipage}[b]{0.09\textwidth}
					test image  \ \vspace{1.0\textwidth} \\
					ground truth \ \vspace{1.0\textwidth} \\
					Segnet		 \ \vspace{1.0\textwidth} \\
					Segnet with post-TV      \ \vspace{0.5\textwidth} \\
					RSegnet      \ \vspace{0.2\textwidth} \\
				\end{minipage}
			}
		\end{subfigure}
		\begin{subfigure}
			{
				\begin{minipage}[b]{0.19\textwidth}
					\includegraphics[width=1\textwidth]{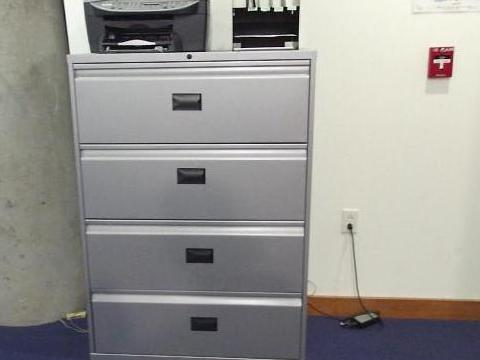} \\
					\includegraphics[width=1\textwidth]{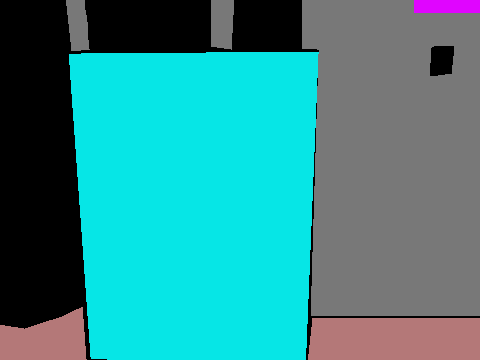} \\
					\includegraphics[width=1\textwidth]{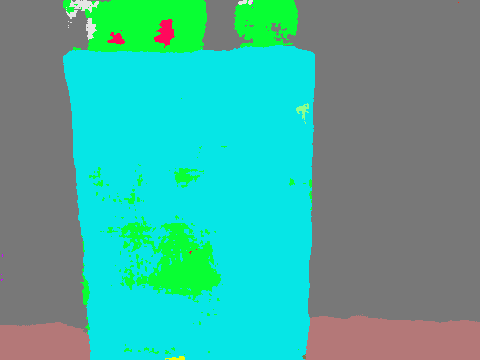} \\
					\includegraphics[width=1\textwidth]{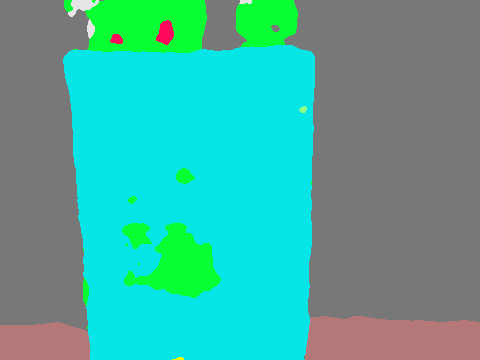} \\
					\includegraphics[width=1\textwidth]{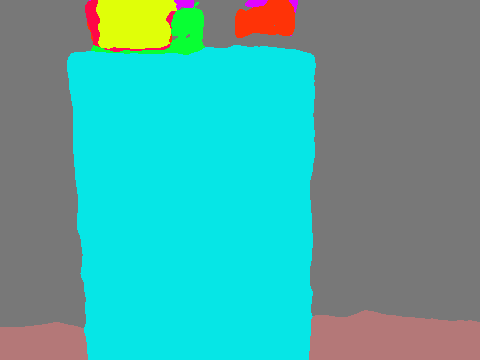}
				\end{minipage}
			}
		\end{subfigure}
		\begin{subfigure}
			{
				\begin{minipage}[b]{0.19\textwidth}
					\includegraphics[width=1\textwidth]{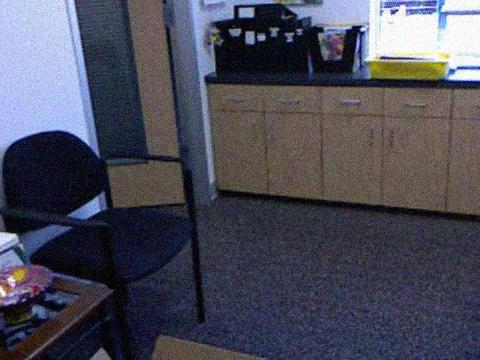} \\
					\includegraphics[width=1\textwidth]{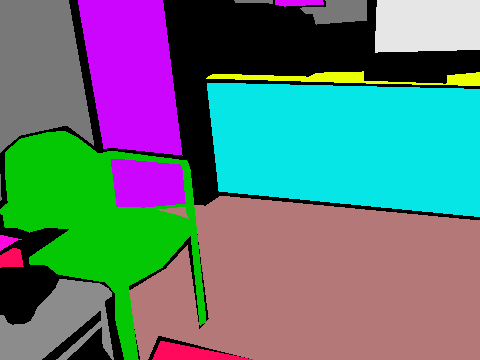} \\
					\includegraphics[width=1\textwidth]{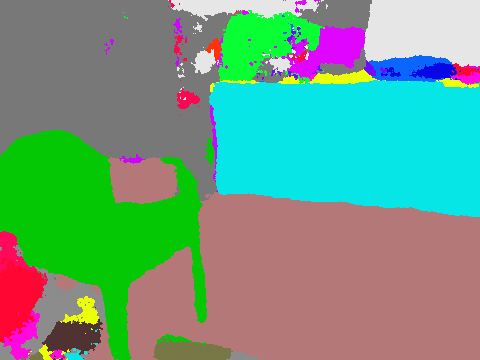} \\
					\includegraphics[width=1\textwidth]{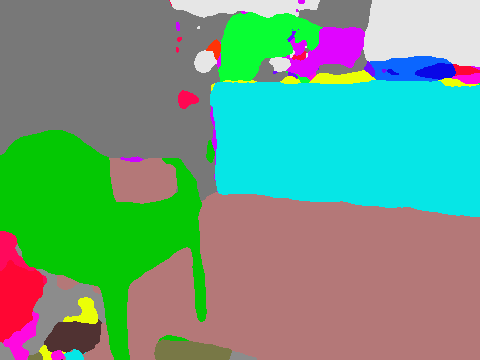} \\
					\includegraphics[width=1\textwidth]{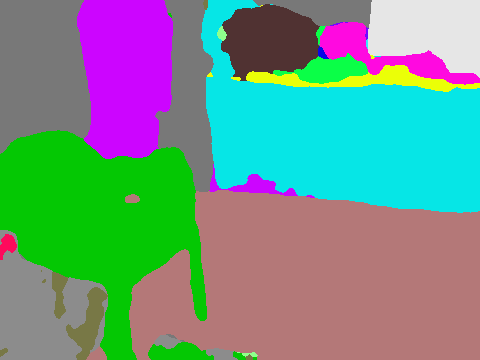}
				\end{minipage}
			}
		\end{subfigure}
		\begin{subfigure}
			{
				\begin{minipage}[b]{0.19\textwidth}
					\includegraphics[width=1\textwidth]{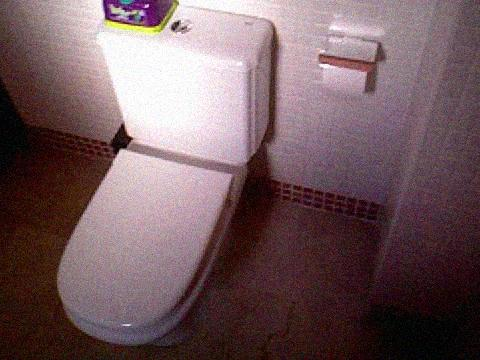} \\
					\includegraphics[width=1\textwidth]{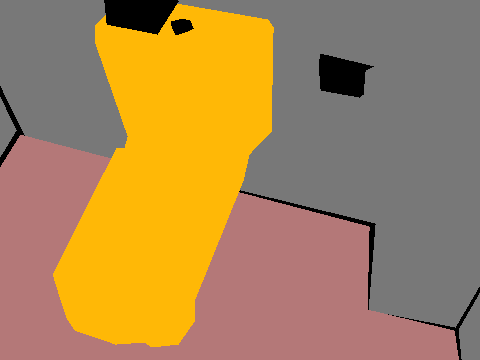} \\
					\includegraphics[width=1\textwidth]{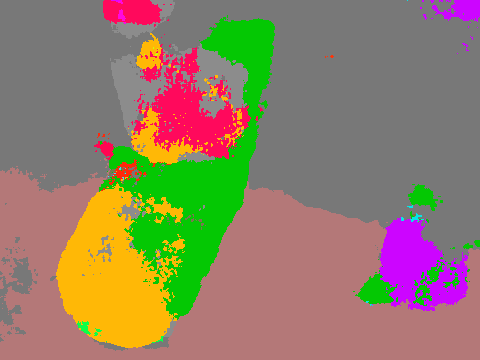} \\
					\includegraphics[width=1\textwidth]{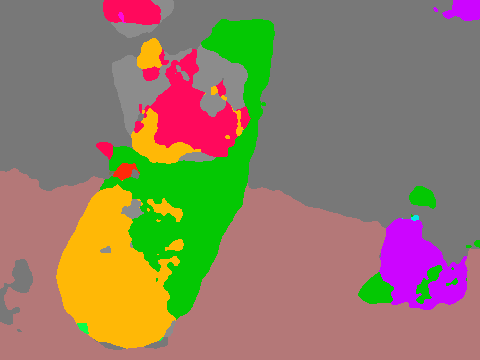} \\
					\includegraphics[width=1\textwidth]{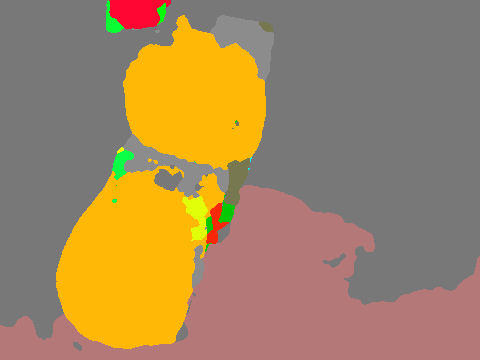}
				\end{minipage}
			}
		\end{subfigure}
		\begin{subfigure}
			{
				\begin{minipage}[b]{0.19\textwidth}
					\includegraphics[width=1\textwidth]{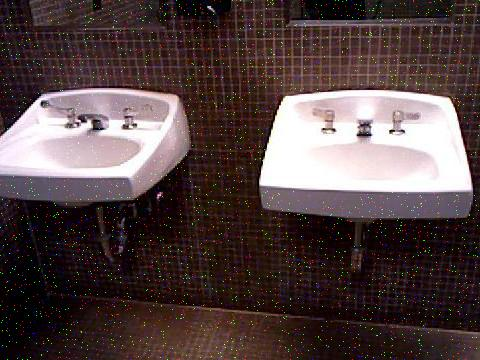} \\
					\includegraphics[width=1\textwidth]{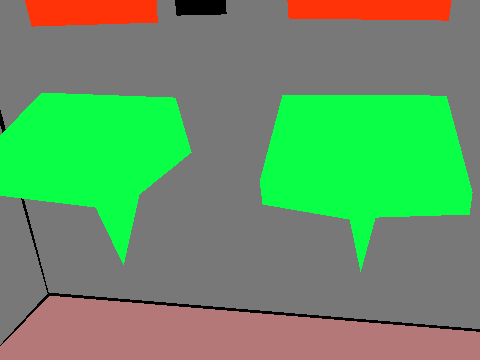} \\
					\includegraphics[width=1\textwidth]{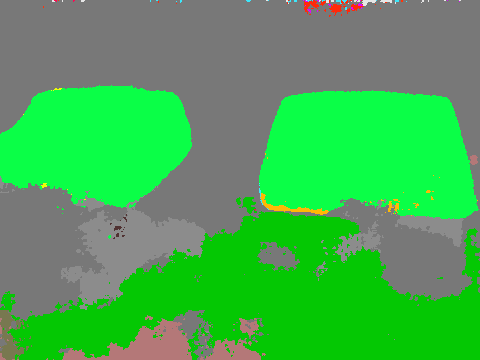} \\
					\includegraphics[width=1\textwidth]{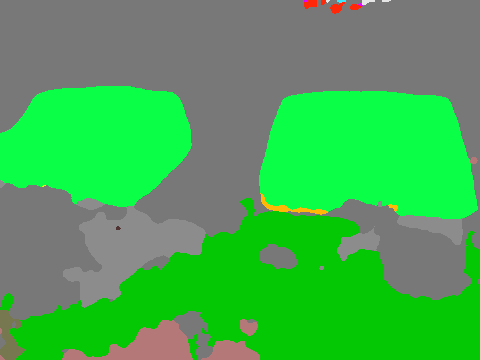} \\
					\includegraphics[width=1\textwidth]{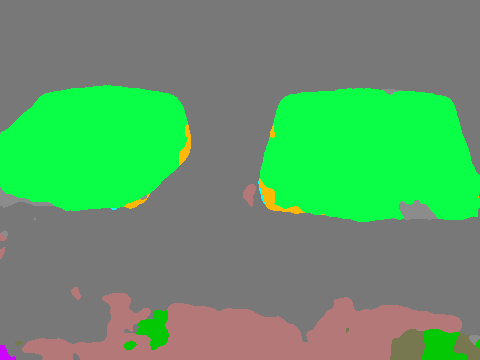}
				\end{minipage}
			}
		\end{subfigure}
		\caption{Segmentation results of Segnet and RSegnet trained on clean dataset. Noise type from left to right: clean image, medium level gaussian noise, medium level gaussian noise, small level salt noise.}
		\label{fig:segmentation_task_6}
	\end{figure*}
	
	We replace original softmax layer with regularized softmax layer, other layers and parameters of Segnet and RSegnet remain the same. Both Segnet and RSegnet are trained for 200k iterations with weights initialized from the VGG model trained on ImageNet, their mini-batch sizes are both three. Learning rates of Segnet and RSegnet are fixed to be 0.001.

	First, Segnet and RSegnet are all trained on clean data and tested on data with different noise.  Then, we randomly pick out 1000 images in training dataset and randomly add
	gaussian noise with zero mean, $\sigma=0.05$ or pepper and salt noise with 1\% pixels' value changed to each image. We make a further comparison when Segnet and RSegnet are trained on noisy dataset. \Cref{table5} and \Cref{table6} show predictions on data with different noise levels.
	
	Since images in SUN-RGBD Dataset are captured by different sensors with different resolutions, quality of images are uneven. Compared to WBC and CamVid datasets, SUN-RGBD dataset is not so clean and tidy.  What's more, thousands labels appear in the original annotation file of SUN-RGBD Dataset, resulting in a much more complex image structure. All these greatly increase the difficulty of the learning task. When introducing the same level noise, the mIoU curves of Segnet and RSegnet are closer than those in the other two datasets due to latent perturbation has been added to SUN-RGBD dataset. Nevertheless, RSegnet still shows better performance than Segnet. It seems that we can benefit more from regularized softmax when it is applied on clean and tidy dataset.

	\section{Conclusions and Future Work}
	\label{sec:conclusions}
	Motivated by the desire for obtaining regularized edges, eliminating scattered points and tiny regions, we propose regularized softmax on CNNs for semantic image segmentation. By applying our method to regularizing Unet and Segnet, we observed better performance from experiments on WBC Dataset, CamVid Dataset and SUN-RGBD Dataset. The proposed method can be applied easily to many other CNNs and tasks. In the future, we will explore the potential to design an end-to-end network by training CNN layers and $\bm \eta$  simultaneously.. 
	
	\FloatBarrier
	\bibliographystyle{siamplain}
	\bibliography{egbib}
	
\end{document}